\documentclass[10pt,twocolumn,letterpaper]{article}

\usepackage[pagenumbers]{cvpr} 

\usepackage{comment}
\usepackage{enumitem}

\usepackage[dvipsnames]{xcolor}
\usepackage{dsfont}
\usepackage{makecell}
\usepackage{units}
\usepackage{multirow} 
\usepackage{tikz}
\usepackage{tabularray}
\usepackage{soul}
\usepackage{multirow}
\usepackage{amssymb}
\usepackage{pifont}

\newcommand{\modelname}{\textsc{TriFuser}\xspace}
\newcommand{\dslname}{\textsc{SplitWeave}\xspace}

\newenvironment{packed_enumerate}{
  \begin{enumerate}[nosep]
}{\end{enumerate}}

\newif\ifshowedits

\newcommand{\addeditor}[3]{%
  \definecolor{#1color}{rgb}{#3}
  \expandafter\newcommand\csname #1\endcsname[1]{%
  \ifshowedits
    {\color{#1color} ##1}%
  \else
    {##1}%
  \fi
  }%
  \expandafter\newcommand\csname #1rmk\endcsname[1]{%
  \ifshowedits
    {\color{#1color} {\bf [#2: ##1]}}
  \fi
  }%
  \expandafter\newcommand\csname #1rpl\endcsname[2]{%
  \ifshowedits
    {{\color{#1color} ##1} \sout{##2}}
  \else
    {##1}
  \fi
  }%
}

\definecolor{uvexprs}{HTML}{DAE8FC}
\definecolor{sexprs}{HTML}{FFF2CC}
\definecolor{spatialcolor}{HTML}{D5E8D4}
\definecolor{utilcolor}{HTML}{CDEB8B}

\newcommand{\mycomment}[1]{}

\definecolor{cbad}{HTML}{EAC2C2}
\definecolor{cmeh}{HTML}{FFE1C9}
\definecolor{cok}{HTML}{FDF3D0}
\definecolor{cgood}{HTML}{B3D09F}

\definecolor{cvprblue}{rgb}{0.21,0.49,0.74}
\usepackage[pagebackref,breaklinks,colorlinks,citecolor=cvprblue]{hyperref}

\def\paperID{1268} 
\def\confName{CVPR}
\def\confYear{2024}

\title{Pattern Analogies: Learning to Perform Programmatic Image Edits by Analogy}

\author{
    Aditya Ganeshan$^*$ \\ 
    Brown University \\ 
    \and
    Thibault Groueix \\ 
    Adobe Research \\ 
    \and
    Paul Guerrero \\ 
    Adobe Research \\ 
    \and
    Radomír Měch \\ 
    Adobe Research \\ 
    \and
    Matthew Fisher \\ 
    Adobe Research \\ 
    \and
    Daniel Ritchie \\ 
    Brown University \\
}

\begin{document}

\definecolor{darkgreen}{RGB}{0,110,0}
\definecolor{darkred}{RGB}{170,0,0}
\def\greencheckmark{\textcolor{darkgreen}{\checkmark}}
\def\redxmark{\textcolor{darkred}{\text{\ding{55}}}}  %

\addeditor{aditya}{AG}{0.7, 0.0, 0.7}
\addeditor{thibault}{TG}{0.0, 0.0, 0.8}
\addeditor{paul}{PG}{0.0, 0.5, 0.0}
\addeditor{radomir}{RM}{0.1, 0.5, 0.9}
\addeditor{matt}{MF}{0.8, 0.0, 0.0}
\addeditor{daniel}{DR}{0.9, 0.2, 0.5}

\showeditsfalse
\showeditstrue

\twocolumn[{%
\renewcommand\twocolumn[1][]{#1}%
\maketitle
\vspace{-2em}
	\centering
	\includegraphics[width=1.0\linewidth]{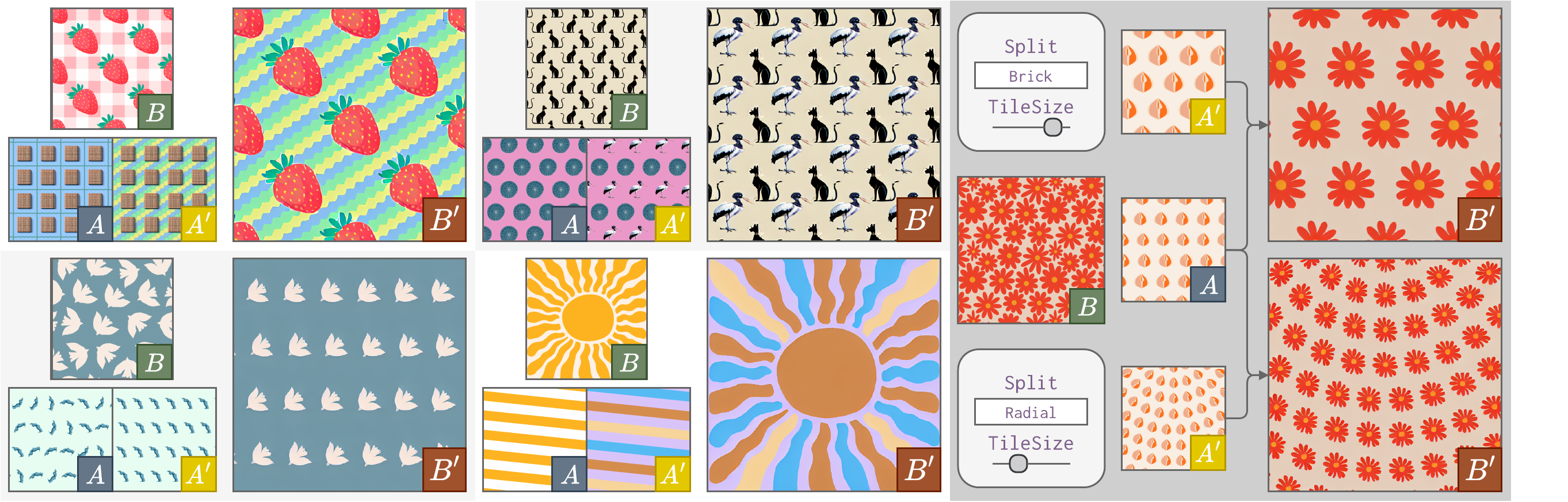}
\captionof{figure}{
Our system performs \textit{programmatic} edits on pattern images without inferring their underlying programs.
(Left) Desired edits, expressed with a pair of patterns $(A, A^\prime)$, are executed on a target pattern $B$ by a generative model to produce $B^\prime$. 
(Right) Parametric changes $A \rightarrow A^\prime$ enabled by our domain-specific pattern language induce corresponding changes to the more complex pattern $B$.  
\vspace{1em}
}
\label{fig:teaser}
}]


\begin{abstract}
Pattern images are everywhere in the digital and physical worlds, and tools to edit them are valuable.
But editing pattern images is tricky: desired edits are often programmatic: structure-aware edits that alter the underlying program which generates the pattern.
One could attempt to infer this underlying program, but current methods for doing so struggle with complex images and produce unorganized programs that make editing tedious.
In this work, we introduce a novel approach to perform programmatic edits on pattern images.
By using a pattern analogy---a pair of simple patterns to demonstrate the intended edit---and a learning-based generative model to execute these edits, our method allows users to intuitively edit patterns.
To enable this paradigm, we introduce \dslname, 
a domain-specific language that, combined with a framework for sampling synthetic pattern analogies, enables the creation of a large, high-quality synthetic training dataset.
We also present \modelname, a Latent Diffusion Model (LDM) designed to overcome critical issues that arise when naively deploying LDMs to this task.
Extensive experiments on real-world, artist-sourced patterns reveals that our method faithfully performs the demonstrated edit while also generalizing to related pattern styles beyond its training distribution.

\end{abstract}

\section{Introduction}
\label{sec:intro}
Visual pattern designs enhance digital media such as presentations, website themes, and user interfaces, and they are woven into the physical world through textiles, wallpapers, and product designs like hardware covers. Given the ubiquity of patterns, methods for editing them are essential: designers should be able to quickly experiment with variations, customize designs to meet specific needs, and adapt existing patterns to align with evolving trends.

\renewcommand{\thefootnote}{\fnsymbol{footnote}}
\addtocounter{footnote}{1}
\footnotetext{Work performed during an internship at Adobe Research}
\renewcommand{\thefootnote}{\arabic{footnote}}

Editing pattern images is not straightforward, as patterns are inherently structured, defined by rules that govern their layout and composition: tiling patterns adhere to principles of alignment and repetition (see Figure~\ref{fig:teaser}: top left), while retro-style designs rely on spatial divisions and fills (see Figure~\ref{fig:teaser}: bottom center).
The edits that designers desire often aim to adjust these underlying organizational rules rather than make superficial, pixel-level changes. We refer to such edits as \textit{programmatic} edits, requiring manipulation of the underlying program that defines a pattern’s structure.

One strategy for enabling such programmatic edits is visual program inference (VPI)~\cite{Mao2019Program, Ellis_NEURIPS2018, acm_match}, where a program that replicates an image is automatically inferred, allowing users to modify the image by adjusting program parameters. However, applying VPI to patterns presents two obstacles.
First, VPI attempts to infer a program that fully replicates a pattern, which can be challenging as patterns are often \textit{semi-parametric}, blending rule-based logic with non-parametric components. 
For instance, the layout of elements in a tiling pattern may be rule-based, but the elements themselves may not be. 
Second, editing with an inferred program can be cumbersome, as they are often poorly-structured, with many unlabeled parameters, making them difficult to interpret. 
Consequently, VPI not only solves a more complex problem than necessary but also makes editing more challenging.

Can we perform programmatic edits without inferring the underlying program?
Doing so requires the ability to \textit{express} and \textit{execute} the edit---both without direct access to the program's parameters. To express a programmatic edit, it’s crucial to specify both \textit{which} underlying parameter(s) to change and \textit{how} to modify them. We draw inspiration from how humans communicate transformations: through analogies. By providing a pair of simple example patterns $(A, A^\prime)$ that illustrate the desired change, users can intuitively convey both aspects of the edit. To execute these edits, we employ a learning-based conditional generative model. Given a pair of simple patterns $(A, A^\prime)$ and a complex target pattern $B$, our system generates $B^\prime$, an edited version of $B$ which performs the transformation demonstrated between $A$ and $A^\prime$ while preserving $B$'s other structural features.
Crucially, $A$ does not need to replicate or even be similar to $B$---it only needs to demonstrate \textit{which} property to edit and \text{how}.
Thus, specifying $A$ is a much easier task than solving VPI.
While prior works~\cite{image_brush, acm_dia, visprompt} have applied analogical editing to image manipulation, they focus primarily on appearance modifications.
In contrast, our approach is the first to use analogies for \textit{programmatic}, structure-aware edits. 
Figure~\ref{fig:teaser} (left) shows examples of analogical editing on complex, real-world patterns.

To make our approach possible, we introduce \dslname: a domain-specific language (DSL) for crafting visual patterns.
\dslname serves two purposes in our method.
First, it enables parametric definition of input pairs $(A, A')$, allowing users to guide transformations in $(B, B')$ as if the underlying program for $B$ were accessible. 
In Figure~\ref{fig:teaser} (right), modifying the \dslname program for $A^\prime$ produces corresponding changes in $B^\prime$. 
Second, \dslname supports the creation of large-scale synthetic training data. We develop program samplers that generate high-quality patterns in two common styles: tiling-based designs with repeating elements and color field patterns characterized by splitting the canvas into intricate colored regions.
Training a model for analogical editing requires a dataset of quartets $(A, A^\prime, B, B^\prime)$.
By applying identical programmatic edits to the \dslname programs for both $A$ and $B$ to produce $A^\prime$ and $B^\prime$, we ensure that the transformation from $A$ to $A^\prime$ mirrors that from $B$ to $B^\prime$. This approach allows us to generate a diverse dataset of analogical quartets.
Models trained on this dataset generalize effectively to real-world patterns within these styles and even extend to related styles.

We use this synthetic dataset to train a novel diffusion-based conditional generative model for executing analogical edits.
Our model directly generates edited patterns $B^\prime$ by conditioning on visual features extracted from input patterns $(A, A^\prime, B)$.
Existing image-conditioned diffusion models~\cite{versatile_diffusion, Rombach_2022_CVPR} prove ineffective, as they fail to interpret the input analogies accurately and neglect fine details.
To address these issues, we incorporate architectural enhancements that enable our model, \modelname, to effectively perform analogical edits. With these improvements, \modelname surpasses prior architectures for analogical editing when applied to pattern images.

To evaluate our method, we curated a test set of $50$ patterns from Adobe Stock spanning $7$ distinct styles. 
A perceptual study on this dataset reveals that participants consistently prefer edits by \modelname over those produced by recent training-free and training-based analogical editing methods.
Despite training only on two of these styles, our model successfully generalizes to the remaining five, which were unseen during training.
On a synthetic validation set with ground-truth analogical edits, our model achieves higher structural and perceptual similarity to the ground truth than prior methods
Finally, we showcase two compelling applications of our approach: mixing attributes of different patterns and transferring pattern animations.

In summary, our contributions are as follows:
\begin{packed_enumerate}
    \item A novel framework for performing \textit{programmatic} edits to pattern images without requiring program inference, leveraging analogies to specify and apply edits.
    \item \dslname, a DSL for crafting a diverse range of visual patterns, designed to support both parametric control and synthetic dataset generation.
    \item A procedure for generating synthetic analogical quartets, enabling editing of \textit{in-the-wild} patterns.
    \item \modelname, a diffusion-based conditional generative model that achieves high fidelity in analogical edits, surpassing prior techniques in both analogical fidelity and generation quality.
\end{packed_enumerate}

\section{Related Work}
\label{sec:related_work}

We review three key areas: (1) Visual Program Inference (VPI) for programmatic editing of structured visual data and its limitations, (2) DSLs and synthetic data generation, specifically for visual patterns, and (3) analogical reasoning in computing, particularly for editing images.
and (4) limitations of alternative editing techniques for patterns, highlighting the need for an analogy-based approach.

\noindent
\textbf{Visual Program Inference for Editing}:
Visual Program Inference (VPI) enables programmatic edits of visual data by inferring executable programs from visual inputs. Prior works have achieved promising results in inferring material graphs~\cite{acm_match, arman_noise, hu2022diff, material_li} and CAD programs for 2D~\cite{Li:2020:Sketch2CAD, NEURIPS2021_2e92962c} and 3D~\cite{extrude_net, zoneGraphs} inputs, using large annotated datasets~\cite{Wu_2021_ICCV, acm_match}, differentiable program approximations~\cite{hu2022diff, CSGSTUMP_ICCV}, or bootstrapped learning~\cite{jones2022PLAD, ganeshan2023coref, jones2024VPIEdit}.
VPI is challenging to adapt to pattern editing due to the scarcity of high-quality annotated pattern data and the non-differentiability of most pattern programs.
Also, VPI approaches often yield complex programs that are difficult to edit and interpret, making them impractical for editing. To address these challenges, recent work has aimed to simplify programmatic editing by inferring edit-specific controls~\cite{parsel, texslider, Cheng_2024_CVPR} or a limited set of semantically meaningful parameters~\cite{jones2021shapeMOD, jones2023ShapeCoder, reparam_cad, sid_yumer}.
Our approach shares this goal of enabling accessible control but extends it further: we transfer control from simple parametric objects to complex in-the-wild images via analogy, bypassing the need for VPI.

\noindent
\textbf{DSL and Synthetic data}
Domain-Specific Languages (DSLs) enable concise descriptions of structured objects, facilitating their creation. Prior works have developed DSLs for Zentangle patterns~\cite{gTangle}, material graphs~\cite{acm_match}, semi-parametric textures~\cite{semi_param_DSL}, and 3D models~\cite{ShapeAssembly, pearl2022geocode}. 
Our DSL focuses on visual patterns constructed through partitioning and merging of canvas fragments. 
Closest to our work is ETD~\cite{element_texture_design_DSL}, which also uses canvas partitioning and merging operators, though it is limited to stationary patterns. 

\noindent
\textbf{Analogical Reasoning}
Analogical reasoning is a foundational AI task: early work includes Evans' ANALOGY program~\cite{evans_old}, CopyCat~\cite{copycat}, and Structure-Mapping Engine~\cite{sem_old}. 
In visual computing, Image Analogies~\cite{aaron_image_analogies} pioneered the concept of analogy-driven editing.
%
Recently, diffusion models have been adapted for analogical editing. DIA~\cite{acm_dia} introduced a training-free approach to analogical editing using pretrained diffusion models. Analogist~\cite{gu2024analogist} offers a complementary method, leveraging inpainting models alongside multimodal reasoning from large language models~\cite{openai2024gpt4}. These training-free approaches are limited to images within the diffusion model’s training domain, limiting their applicability to patterns. Other methods attempt to learn analogical editors by finetuning diffusion models on analogical pairs~\cite{image_brush, meng2024instructgie, visprompt}. 
However, the focus of all these works remains largely on stylistic, appearance edits, often failing to perform \textit{programmatic} edits. This limitation arises both from the models’ architectures and from the lack of training pairs with \textit{programmatic} edits. Our work addresses both these gaps, enabling structured, programmatic analogical edits for visual patterns.

\begin{figure}[t!]
	\centering
	\includegraphics[width=1.00\linewidth]{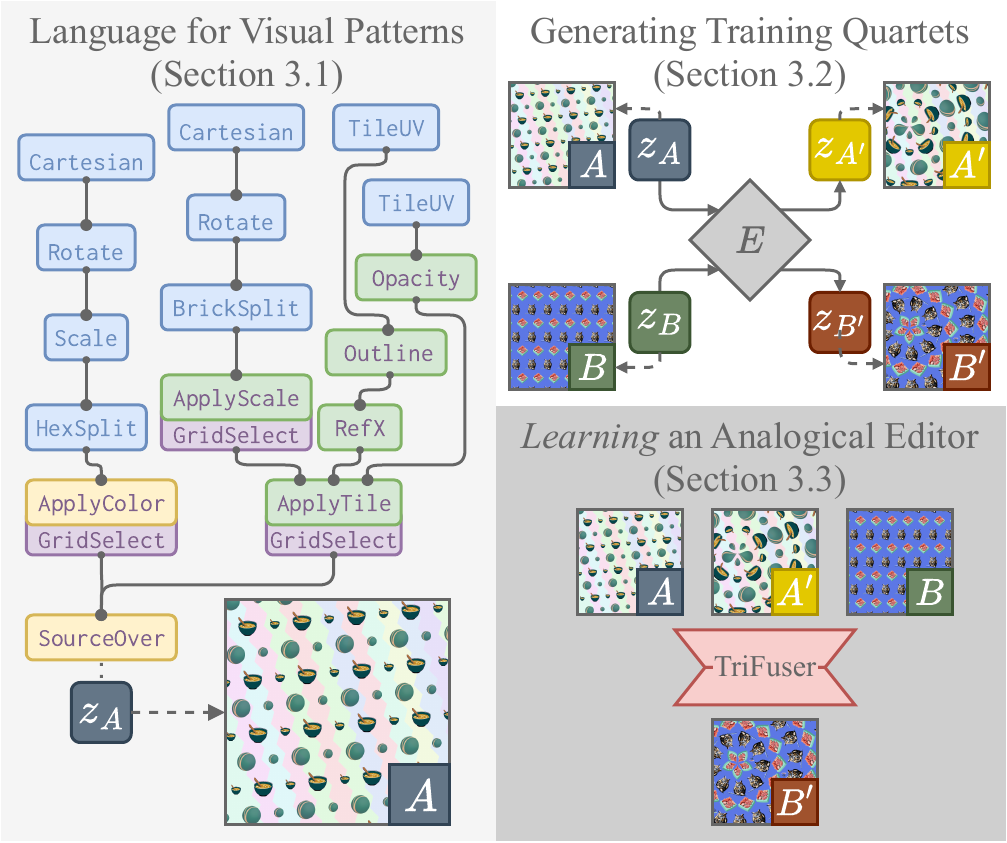}
	\caption{ \textbf{Overview}: To create high-quality visual patterns, we introduce a custom DSL called \dslname. Pairs of \dslname programs $(A, B)$ are then jointly edited to create analogical quartets. This synthetic data is then used to train \modelname, a neural network for analogical pattern editing.
 }
\label{fig:overview}
\end{figure}

\noindent
\textbf{Alternate approaches}
Many works propose to edit images via style transfer~\cite{gatys_style, analogy_style}, image-to-image translation~\cite{CycleGAN2017, image_translateion_eccv24}, or remixing image decompositions~\cite{BKPB17, yotam_image_vectorization, mu2024editable}. These methods typically separate an image’s appearance and structure, allowing the appearance to be replaced or modified from external sources, but they are limited to stylistic edits and are not suitable for programmatic transformations.
More recent methods manipulate cross-attention maps~\cite{cross_edit} or prompt tokens~\cite{hertz2022prompt} in diffusion models to achieve specific image edits, with approaches often tailored to particular transformations, such as geometric changes~\cite{sajnani2024geodiffuser, loose_control} or stylistic adjustments~\cite{wallace2022edict, hertz2022prompt}. In contrast, exemplar pairs offer a flexible way to demonstrate a variety of edits without requiring specialized models for each type.


\section{Method}
\label{sec:method}

Our objective is to enable programmatic edits of 2D visual patterns without inferring their underlying programs. Instead, we propose an alternative that uses analogies to \textit{express} desired edits and a conditional generative model to \textit{execute} them.
Formally, given two source patterns $A$ and $A^\prime$ that demonstrate a desired edit, along with a target pattern $B$, our goal is to generate an edited target pattern $B^\prime$ that applies this edit to $B$. This task is defined as learning a mapping $f(A, A^\prime, B) \rightarrow B^\prime$, where $A$, $A^\prime$, $B$, and $B^\prime$ are 2D RGB images ($\in \mathbb{R}^{H \times W \times 3}$).
To learn this mapping, we generate a large synthetic dataset of analogical pattern quartets $(A, A^\prime, B, B^\prime)$. 

Figure~\ref{fig:overview} provides a schematic overview of our approach. 
First, in Section~\ref{subsec:dsl} we introduce \dslname, a Domain-Specific Language (DSL) that enables the creation and manipulation of various kinds of patterns. 
Section~\ref{subsec:quartets} describes our approach for sampling analogical quartets in \dslname to create the synthetic training data. 
Finally, in Section~\ref{subsec:model}, we present \modelname, a conditional generative model that learns to \textit{execute} analogical edits.

\subsection{A Language for Visual Patterns}
\label{subsec:dsl}

To enable programmatic edits without program inference, our approach requires two core capabilities: (a) generating a large, high-quality synthetic dataset essential for training models to reliably \textit{execute} analogical edits, and (b) the ability to create and parametrically control analogy inputs at test time to effectively \textit{express} desired edits. Existing pattern generation tools are insufficient for these needs, as they are either limited to narrow pattern domains~\cite{gTangle} or demand intense coding effort to produce diverse, high-quality patterns~\cite{element_texture_design_DSL, mccarthy2015getting}. To address these limitations, we introduce \dslname, a DSL designed specifically to support analogical transformations in visual patterns. \dslname combines abstractions for pattern synthesis with a node-based visual programming interface (see Supplementary), enabling efficient generation of high-quality synthetic patterns for training while allowing flexible, precise pattern manipulation to define analogy inputs at test time.

\begin{figure}[t!]
	\centering
	\includegraphics[width=1.00\linewidth]{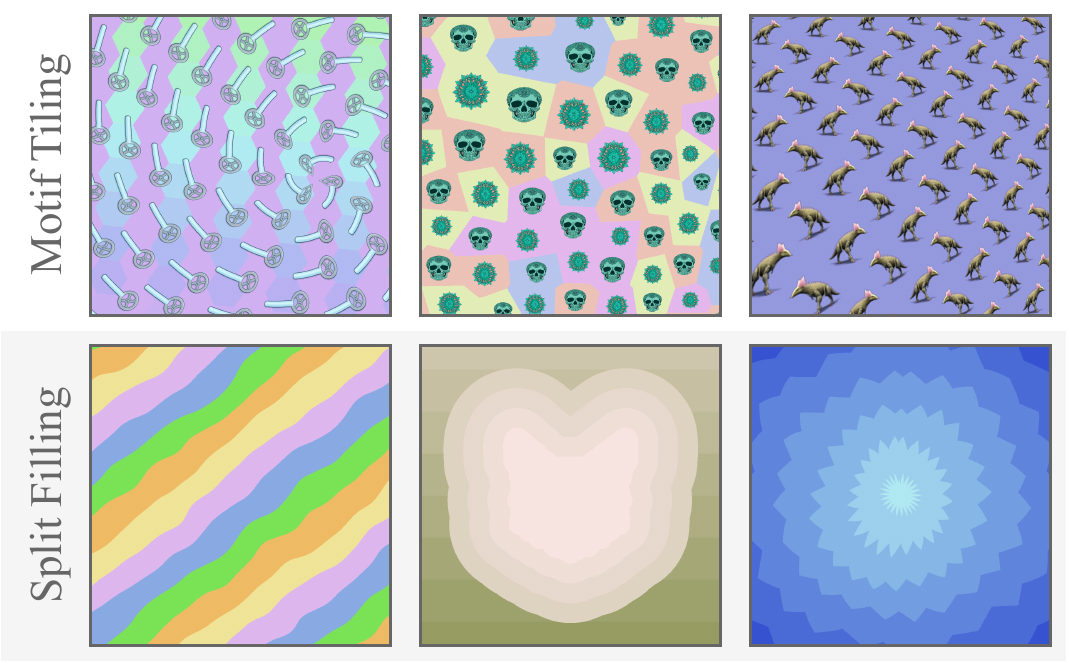}
	\caption{ 
    Custom program samplers for two pattern styles. Our samplers produce diverse and high-quality patterns, enabling generalization to real-world patterns.
 }
\label{fig:synthetic_patterns}
\end{figure}
\dslname uses three types of operations for structured pattern creation: (1) \textit{Canvas Fragmentation}, which allows structured divisions of the canvas, such as brick-like or voronoi splits; (2) \textit{Fragment ID-Aware Operations}, enabling transformations that vary across fragments (e.g., scaling alternating rows or columns) to support spatial variability in non-stationary pattern designs; and (3) various \textit{SVG Operators} for outlining, coloring, and compositing. Together, these operations enable efficient creation of patterns with complex structure and visual variety. Figure~\ref{fig:overview} (left) illustrates these capabilities in a \dslname program for generating a tiling pattern design.

Our goal is to generate high-quality synthetic patterns using \dslname that enable trained editing models to generalize well to real-world patterns. Naive sampling from the DSL grammar often leads to overly complex or incoherent patterns, limiting their effectiveness in model training. Instead, we draw inspiration from recent advances in fields such as geometric problem solving~\cite{alphageometry} and abstract reasoning~\cite{ellis_arc}, where tailored data generators have proven essential for tackling complex tasks. Following a similar approach, we design custom program samplers for two versatile and widely-used pattern styles.
The first, \textit{Motif Tiling Patterns (MTP)}, consists of compositions based on repeated \textit{Tile} elements. These patterns exhibit controlled variations in tile properties across the canvas (e.g. orientation, color, and scale), creating visually cohesive yet richly diverse structures. The second, \textit{Split-Filling Patterns (SFP)}, are generated by dividing the canvas into ordered fragments, applying region-specific coloring and transformations based on fragment IDs. Both pattern styles are common in digital design and support a wide range of programmatic variations, making them particularly suited for analogical editing tasks. Example patterns generated by our program samplers are shown in Figure~\ref{fig:synthetic_patterns}; additional implementation details are in the supplementary materials.

\begin{figure}[t!]
	\centering
	\includegraphics[width=1.00\linewidth]{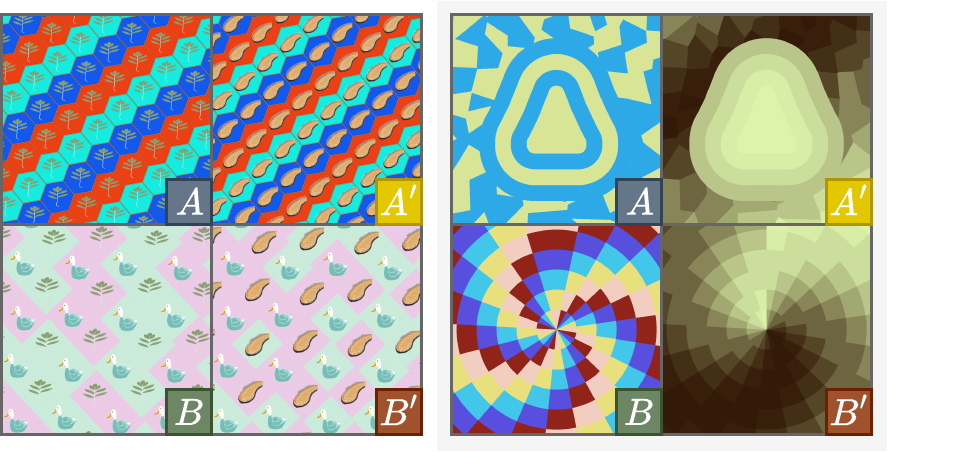}
	\caption{ 
    We create synthetic analogical quartets $(A, A^\prime, B, B^\prime)$ with consistent edits between $A$ and $B$ pairs, providing  data for training an analogical editing models.
 }
\label{fig:synthetic_analogies}
\end{figure}

\begin{figure*}[t!]
	\centering
	\includegraphics[width=1.00\linewidth]{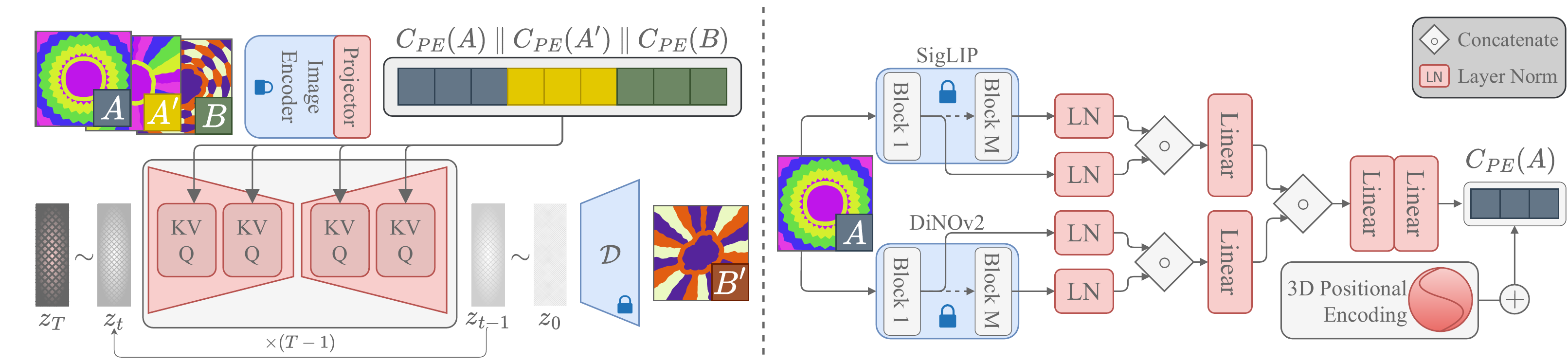}
	\caption{ 
       (Left) \modelname is a latent diffusion model conditioned on patch-wise tokens of the input images $(A, A^\prime, B)$ to generate the analogically edited pattern $B^\prime$. 
(Right) To achieve high-quality edits, we enrich these tokens by fusing multi-level features from multiple encoders, followed by a 3D positional encoding: 2D to specify spatial locations and 1D to specify the token's source ($A$, $A^\prime$ or $B$).
 }
\label{fig:model-overview}
\end{figure*}
\subsection{Sampling Analogical Quartets}
\label{subsec:quartets}
With the ability to generate diverse synthetic patterns using \dslname\ (Section~\ref{subsec:dsl}), our goal is now to construct analogical pattern quartets $(A, A^\prime, B, B^\prime)$. 
Each pattern image in a quartet is generated by a \dslname program $z$. 
These quartets serve as structured training data for editing models, allowing them to learn consistent transformations that can generalize across different pattern domains.

Analogies in our framework are grounded in Structure Mapping Theory~\cite{SMT_dedre}, which defines analogies as mappings of relational structure from a base to a target domain. We designate $(A, A^\prime)$ as the base and $(B, B^\prime)$ as the target, with the requirement that the relationship $R$ between program pairs $(z_A, z_{A^\prime})$ and $(z_B, z_{B^\prime})$ remains consistent:
\begin{equation}
\label{eq:analogy}
R(z_A, z_{A^\prime}) = R(z_B, z_{B^\prime}).
\end{equation}
Rather than focusing on visual similarity between the patterns $(A, A^\prime)$ themselves, this program-level analogy allows us to generate quartets with transformations that affect the underlying program, facilitating \textit{programmatic} edits.

To construct these analogical quartets, we use a program sampler along with a predefined set of editing operators $E$. For each quartet, we begin by sampling an edit $e \in E$, followed by sampling initial programs $z_A$ and $z_B$ that are compatible with $e$.
Applying $e$ to both $z_A$ and $z_B$ yields transformed programs $z_{A^\prime}$ and $z_{B^\prime}$. By using identical transformations across domains, we ensure a consistent ``edit relation'' across the quartet, satisfying Equation~\ref{eq:analogy} by construction.
In Figure~\ref{fig:synthetic_analogies}, we illustrate examples of synthetic analogical quartets generated using this method, demonstrating consistent transformations between $(A, A^\prime)$ and $(B, B^\prime)$.

\noindent
\textbf{Edit Operators $E$.} We focus on edits targeting specific sub-parts of the program. Specifically, we consider three types of edits: \textit{insertion}, \textit{removal}, and \textit{replacement} of sub-programs. For example, an edit operator might \textit{replace} the sub-program responsible for splitting the canvas, while other edits may \textit{insert} or \textit{remove} tiles within the pattern. Please refer to the supplementary for more details.

\subsection{Learning an Analogical Editor}
\label{subsec:model}

Our goal is to train a model on the synthetic data that is capable of performing analogical edits on real, \textit{in-the-wild} patterns. 
Specifically, we aim to generate the target pattern $B^\prime$ from an input triplet $(A, A^\prime, B)$.
This approach allows users to demonstrate desired edits with a simple pattern pairs $(A, A^\prime)$, which the model then applies to a complex patterns $B$ to produce $B^\prime$.
Given the success of Latent Diffusion Models (LDMs) in various generative modeling tasks~\cite{Rombach_2022_CVPR}, we chose to adapt an LDM for our task as well. 
We propose \modelname, a latent diffusion model (LDM) for analogical editing (Figure~\ref{fig:model-overview}).
We provide a brief overview of LDMs to provide context before detailing \modelname’s modifications for analogical editing.

\noindent
\textbf{Preliminaries}: 
Denoising Diffusion Probabilistic Models (DDPMs)~\cite{ho2020denoising} transform random noise into structured data via reverse diffusion steps guided with a conditioning embedding $c(y)$ (often derived from text).
Latent Diffusion Models (LDMs) extend DDPMs by mapping data to a lower-dimensional latent space via an encoder. During training, a UNet model~\cite{unet} learns to remove noise introduced into the latents. 
During inference, a latent sampled from a normal distribution is iteratively denoised by the model to yield a clean latent.
Finally, the clean latent is decoded to generate the output image.
Please refer to~\cite{yang2023diffusurvey} for a more thorough overview. 
For analogical editing we adapt an Image Variation (IM) model~\cite{versatile_diffusion}, which uses patch-wise image tokens extracted using a text-image encoder~\cite{pmlr-v139-radford21a} as the conditioning embedding $c(y)$.


The simplest adaptation of an IM model to our task is to generate $B^\prime$ conditioned on image tokens from all three input images, concatenated as $C = c(A) \mathbin\Vert c(A^\prime) \mathbin\Vert c(B)$, where $\mathbin\Vert$ denotes token-wise concatenation. This approach, however, suffers from three drawbacks: \textit{Token Entanglement}, \textit{Semantic Bias}, and \textit{Detail Erosion}. We discuss each of these issues briefly, along with our solutions.

\noindent
\textbf{Detail Erosion}: Despite using patch-wise tokens, the extracted features lack the fine-grained information needed to retain key aspects of $B$ in the generated pattern $B^\prime$. Consequently, the model often struggles to preserve elements like tile textures. To address this problem, we combine features from both the first and last layers of the feature encoder:
\begin{equation}
C_{\textit{hl}}(P) = \textit{Linear}(\textit{LN}(c_{\textit{high}}(P)) \cdot \textit{LN}(c_{\textit{low}}(P)),
\end{equation}
where $\textit{LN}$ is layer normalization, $\cdot$ denotes channel-wise concatenation, $P$ is an input pattern, and $\textit{Linear}$ is a linear projection layer that fuses low- and high-level features.

\noindent
\textbf{Semantic Bias:} Image variation models typically use feature extractors such as CLIP~\cite{pmlr-v139-radford21a}, which are trained to align image embeddings with corresponding text embeddings.
Such embeddings emphasize high-level semantics but lack spatial and fine-grained visual details. Combining these embeddings with features from text-free, self-supervised extractors, such as DiNO~\cite{caron2021emerging}, has been shown to improve performance in downstream tasks~\cite{tong2024eyes, jiang2024from}. For our task, a similar approach---combining features from both text-image ($m_1$) and self-supervised ($m_2$) feature extractors---significantly enhances generation quality.
The extracted features are fused as follows:
\begin{equation}
 C_{\textit{mix}}(P) = \textit{Mixer}(C^{m_1}_{\textit{hl}}(P) \cdot C^{m_2}_{\textit{hl}}(P)),
\end{equation}
where $\textit{Mixer}$ is a two-layer MLP that integrates features from the two extractors.

\noindent
\textbf{Token Entanglement}: To successfully perform an analogical edit, for each patch-level feature token, the model must be able to identify to which source image ($A$, $A^\prime$, or $B$) that patch belongs as well as the 2D position of the patch within that image.
Without these distinctions, the model often fails to identify the pattern to edit (i.e., $B$) and to recognize the desired edit from $(A, A^\prime)$. To address this problem, we introduce 3D positional encodings: two dimensions for spatial location within each pattern and one dimension for the source image. These encodings are applied to the extracted embeddings, yielding:
\begin{align}
C^\Omega &= C_{\textit{PE}}(A) \mathbin\Vert C_{\textit{PE}}(A^\prime) \mathbin\Vert C_{\textit{PE}}(B),\\
 C_{\textit{PE}}(P)^{xy} &= C_{\textit{mix}}(P)^{xy} + \textit{PE}(t_P, x, y),
\end{align}
where $t_P$ is a one-hot vector encoding which input image a token comes from and $\textit{PE}(t_P, x, y)$ positionally encodes both spatial and source information for each token.

As we demonstrate in Section~\ref{subsec:ablation}, conditioning on $C^{\Omega}$ instead of $C$ significantly enhances the quality of patterns generated by \modelname. Our adapted architecture, shown in Figure~\ref{fig:model-overview}, integrates the modifications described above to effectively address the described drawbacks. 
Note that this architecture is designed for general analogical editing, not just pattern editing.
To enhance generalizability to real-world patterns, we initialize \modelname with an existing pretrained IM model~\cite{versatile_diffusion}, and fine-tune only the denoising UNet and the projection layers in our feature extractor.

\section{Experiment}
\label{sec:experiments}

In this section, we evaluate our approach along three directions: (1) the effectiveness of \modelname at performing analogical edits on complex, real-world patterns, emphasizing how our synthetic data enables editing of in-the-wild pattern images; (2) the ability of \modelname to support \textit{programmatic}, structure-preserving edits without explicit program inference; and (3) the impact of architectural modifications introduced in \modelname on the quality of generated patterns.
We conduct a human perceptual study, quantitative assessments, and qualitative comparisons to demonstrate our system's ability to perform high-quality analogical edits across a range of pattern types.

\begin{figure}[t!]
	\centering
	\includegraphics[width=1.0\linewidth]{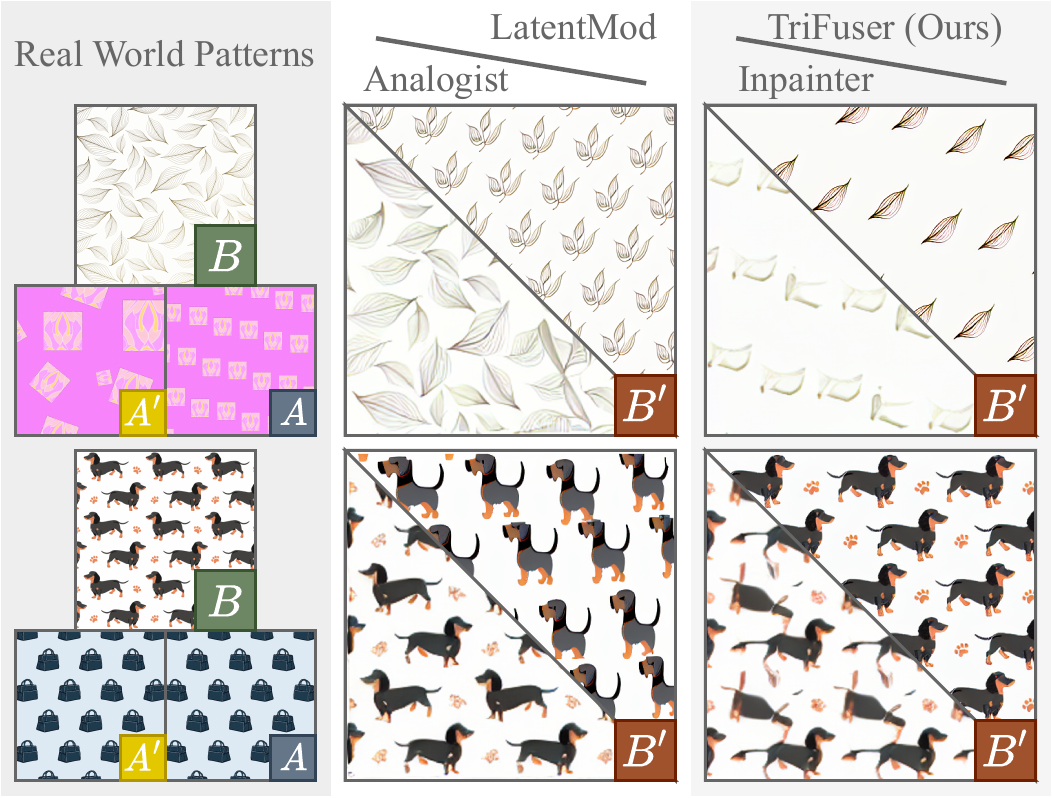}
	\caption{
    Qualitative comparison between patterns generated by our model, \modelname, and the baselines. 
    \modelname generates higher quality patterns with greater fidelity to the input analogy.
 }
\label{fig:qualitative_comparison}
\end{figure}

\subsection{Experiment Design}
\label{subsec:exp_details}
\noindent
\textbf{Datasets}:
We generate a large synthetic dataset of analogical quartets, i.e., pairs of analogical patterns $(A, A^\prime, B, B^\prime)$, using the \dslname program samplers introduced in Section~\ref{subsec:dsl}. 
This synthetic dataset contains approximately 1 million samples covering two pattern styles, namely Split Filling Patterns (SFP) and Motif Tiling Patterns (MTP) (cf. Section~\ref{subsec:dsl}).
For MTP patterns, we synthesize 100k distinct tiles using the LayerDiffuse~\cite{Zhang2024} model, guided by text prompts derived from WordNet~\cite{miller-1994-wordnet} noun synsets. Additionally, we construct a synthetic test set with 1000 analogical quartets to evaluate model performance on unseen synthetic data. Further details on dataset construction are provided in the supplementary material.

To assess \modelname on real-world patterns, we curate a test dataset of 50 patterns created by professional artists and sourced from Adobe Stock. 
This dataset spans seven distinct sub-domains of 2D patterns, representing a range of pattern styles.
These styles include MTP and SFP patterns as well as previously unseen pattern styles such as Memphis-style, geometric, and digital textile patterns. 
Each pattern is annotated with a desired edit, and we use \dslname to generate a pair of simpler patterns $(A, A^\prime)$ demonstrating this edit.
This test set provides a challenging benchmark to evaluate \modelname's generalization to diverse, real-world editing tasks.

\noindent
\textbf{Training details}: We fine-tune a pre-trained diffusion model using our synthetic dataset of analogical quartets, as described in the previous section. We initialize our model with Versatile-Diffusion's Image Variation model~\cite{versatile_diffusion}. 
We use SigLIP~\cite{Zhai_2023_ICCV} as our text-image feature encoder and DiNOv2~\cite{dino_v2} for self-supervised features.
We fine-tune the model on $8$ A100 GPUs using a batch size of $224$ for $\sim65$ epochs over $7$ days. 
During inference, we generate each edited pattern $B^\prime$ with typical diffusion parameter settings such as a classifier-free guidance weight of $7.5$ and $50$ denoising steps.

\subsection{Analogical Editing Baselines}

To evaluate the analogical editing capability of \modelname, we compare it to three baseline methods, each representing a leading approach for analogical image editing.

First, we consider \textit{training-free editors} and \textit{latent arithmetic editors}. Training-free editors repurpose pre-trained diffusion models to perform analogical edits without additional training~\cite{acm_dia, gu2024analogist}, leveraging the rich representations learned by diffusion models for editing. In this category, we compare against \textit{Analogist}~\cite{gu2024analogist}, the current state-of-the-art method. Latent arithmetic editors, on the other hand, rely on transformations in a learned latent space to infer analogical modifications~\cite{analogy_neurips15, tewel2021zero}. Note that these approaches only require samples from the target domain, not analogical training pairs. We implement a baseline for this method by fine-tuning a naive Image Variation model~\cite{versatile_diffusion} on our synthetic dataset to learn a generative latent embedding space of patterns. At inference, analogical edits are generated using latent arithmetic: given patterns $A$, $A'$, and $B$, we condition the generation of $B'$ on $E(B) + E(A') - E(A)$. We refer to this baseline as \textit{LatentMod}.

\newcommand{\newcheckmark}{\raisebox{0.6ex}{\scalebox{0.7}{$\sqrt{}$}}}
\newcommand{\newcrossmark}{\scalebox{0.85}[1]{$\times$}}

\begin{table}[t!]
    \centering
    \begin{tabular}{lc}
        \toprule
        & Preference Rate\\
        \midrule
         \modelname vs. \textit{Analogist} & $89.24\%$ \\
         \modelname vs. \textit{LatentMod} & $80.59\%$ \\
         \modelname vs. \textit{Inpainter} & $72.05\%$ \\
        \bottomrule
    \end{tabular}
    \caption{Results of a two-alternative forced-choice perceptual study comparing our model (\modelname) against three baselines.
    Ours is preferred in the overwhelming majority of judgments.
    }
    \label{tab:perceptual_study}
\end{table}

Finally, we consider \textit{analogy-conditioned generative editors}, where models are explicitly trained on analogical data to learn analogical transformations~\cite{image_brush}. This category includes our proposed \modelname as well. 
Image Brush, the state-of-the-art method, fine-tunes a diffusion inpainting model for analogical editing with multi-modal conditioning. 
Since code for Image Brush is unavailable, we implement a similar baseline by fine-tuning a Stable Diffusion inpainting model. This model, which we term \textit{Inpainter}, performs analogical editing by inpainting the lower-left quadrant of a 2x2 analogy grid containing $(A, A^\prime, B)$ and conditioned on a fixed text template. 

\subsection{Editing Real-World Patterns}

To evaluate \modelname's real-world analogical editing capabilities, we conducted a human preference study on the curated test set of Adobe Stock patterns.

We performed a two-alternative forced-choice perceptual study comparing \modelname with baseline methods on all 50 entries in the test set.
Each method generates $k=9$ outputs for each input tuple, and we select the best one based on visual inspection.
Participants were shown edited patterns generated by two different methods along with the input patterns $(A, A^\prime, B)$ and instructed to select the edit that best preserved the analogical relationship and exhibited higher image quality. 
We recruited 42 participants for the study, resulting in a total of 1720 total judgments.

Table~\ref{tab:perceptual_study} presents the results, showing that \modelname was preferred over both \textit{Analogist} and \textit{LatentMod}. Due to the domain gap between the training data of the underlying model~\cite{Rombach_2022_CVPR} and pattern images, \textit{Analogist} fails to interpret and edit pattern images. 
Meanwhile, \textit{LatentMod} fails to perform reasonable edits as the embedding space lacks the low-level details necessary for \textit{programmatic} edits
While these baselines perform adequately on stylistic edits, they are unsuitable for \textit{programmatic} editing.
When compared to \textit{Inpainter}, \modelname was favored in 72.05\% of comparisons. 
Both methods benefit from training on analogical quartets, yet \textit{Inpainter} sacrifices pattern quality as it generates the edited pattern in only a quarter of the full canvas resolution.

\begin{figure}[t!]
	\centering
	\includegraphics[width=1.00\linewidth]{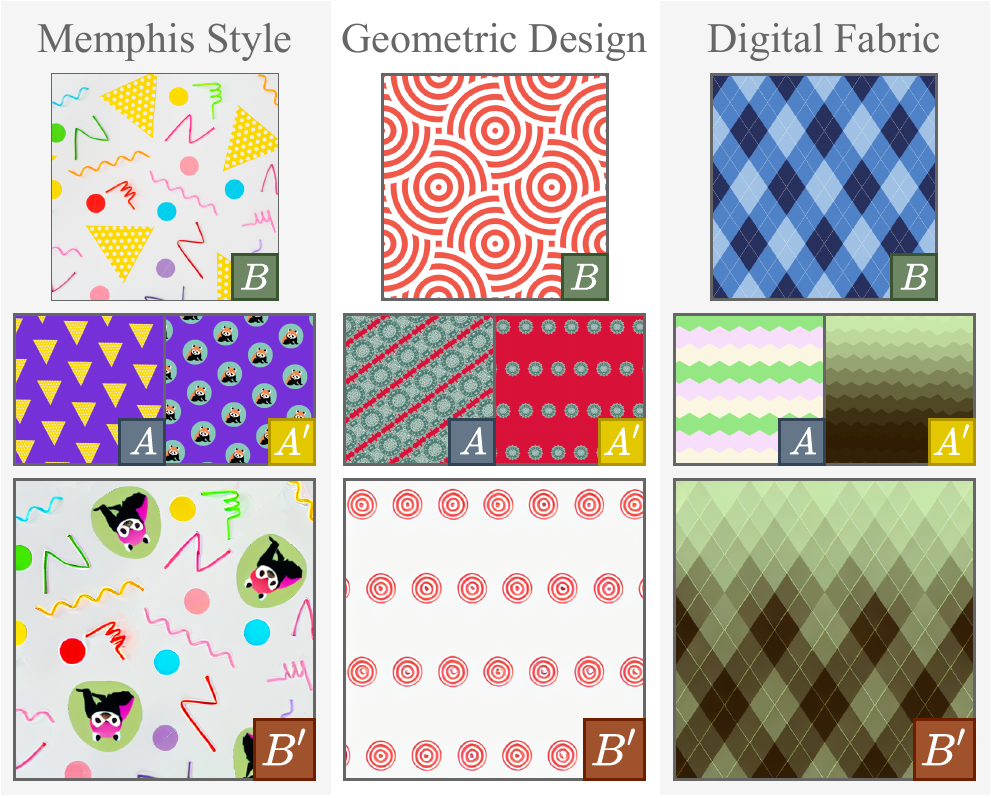}
	\caption{ 
    \modelname effectively edits patterns from novel pattern styles not present in the training dataset. \modelname shows a noteworthy ability to generalize beyond its training distribution.
 }
\label{fig:ood}
\end{figure}

Figure~\ref{fig:qualitative_comparison} shows examples of pattern edits generated by \modelname and the baselines, with our model consistently delivering superior results.
In Figure~\ref{fig:ood}, we show examples of \modelname’s edits on out-of-distribution pattern styles not present in the training set. These results suggest that our synthetic training data enables manipulation of real-world patterns, even extending to certain untrained pattern styles.

\subsection{Editing Synthetic Patterns}
\label{subsec:ablation}

Next, we evaluate \modelname's ability to perform \textit{programmatic} edits on the synthetic validation set, which contains ground truth patterns $B^{\prime}$. Ideally, this would involve verifying that the underlying program $z_{\hat{B}^{\prime}}$ of the generated pattern reflects the same transformation from $z_B$ as that between $z_A$ and $z_{A}^{\prime}$. However, this would require visual program inference on $\hat{B}^{\prime}$, which is infeasible. Instead, we approximate this criterion by comparing the program outputs $\hat{B}^{\prime}$ and $B^{\prime}$ to see if the visual results align with the intended transformation.
To quantify this alignment, we use perceptual metrics—DSim~\cite{fu2024dreamsim}, DIST~\cite{ding2020iqa} and LPIPS~\cite{zhang2018perceptual}—along with SSIM to capture pixel-level structural similarity.

Note that analogies can have multiple valid interpretations, and even a single interpretation may yield several visually-related variations. To account for this multiplicity, we generate $k=5$ output patterns for each input set $(A, A^{\prime}, B)$ and select the one that maximizes each metric. In other words, we evaluate whether at least one generated output aligns with the intended target.

\begin{table}[t]
\small
\begin{center}
\begin{tabular}{lcccc}
\toprule
& {\footnotesize DSIM ($\downarrow$)} & {\footnotesize DISTS ($\downarrow$)} & {\footnotesize LPIPS ($\downarrow$)}& {\footnotesize SSIM ($\uparrow$)}\\
\midrule
{ \textit{Analogist}} & 0.496 & 0.432 & 0.697 & 0.494\\
{ \textit{LatentMod}} & 0.242 & 0.320 & 0.613 & 0.502\\
{ \textit{Inpainter}} & 0.092 & 0.256 & 0.371 & \textbf{0.713}\\
\midrule
{ \modelname}& \textbf{0.074} & \textbf{0.184} & \textbf{0.304} & 0.704 \\
\bottomrule
\end{tabular}
\end{center}

\caption{
Quantitative evaluation on the synthetic validation set shows that \modelname generates patterns with higher perceptual similarity to the ground truth than the baselines.
}
\label{tab:val_eval}
\end{table}

Table~\ref{tab:val_eval} shows the results of this experiment. First, we note that \modelname outperforms all baselines across all perceptual metrics. These metrics capture different aspects of perceptual similarity~\cite{fu2024dreamsim}, and superior performance across all of them suggests a comprehensive improvement.
Second, we observe that the analogy-conditioned generative editors (\textit{Inpainter} \& \modelname) surpass both the training-free and latent modification editors. Interestingly, \textit{Inpainter} achieves slightly higher SSIM scores than \modelname, suggesting that future methods combining elements of both models might be fruitful.

\subsection{\textbf{\modelname} Ablation}

To evaluate the contributions of each model component introduced in Section~\ref{subsec:model}, we conduct a subtractive analysis on the synthetic validation set, using the same perceptual and structural metrics as above. 
For this ablation study, we remove each component one at a time and measure the resulting performance, as reported in Table~\ref{tab:model_ablation}. The results demonstrate that removing any single modification leads to a performance drop, with the removal of 3D positional encoding causing the most severe degradation. This is understandable: without 3D positional encoding, the network often fails to accurately identify which pattern to edit.
For comparison, we also include results from the original Image Variation model~\cite{versatile_diffusion} trained without any modifications (Base). As expected, this model performs poorly, underscoring the importance of our modifications in achieving high-quality analogical edits.

\section{Application}

\begin{figure}[t!]
	\centering
	\includegraphics[width=1.00\linewidth]{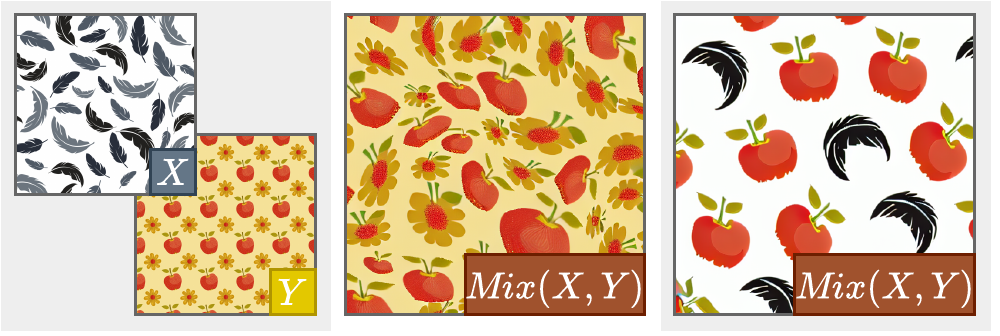}
	\caption{ 
    Our model helps users mix elements of different real-world patterns together, accelerating design exploration.
 }
\label{fig:pattern_mixing}
\end{figure}

The ability to edit patterns without requiring program inference unlocks new creative possibilities. We demonstrate two practical applications of analogical pattern editing:

\noindent
\textbf{Pattern Mixing}:
Figure~\ref{fig:pattern_mixing} shows example of using our method to \emph{mix} elements of two real-world patterns $X$ and $Y$, allowing the user to create unique, hybrid designs.
The \textit{Mix} operator is implemented by using a synthetic pair $(A,A^\prime)$ to create a variant $X^\prime$ of $X$ and then using the pair $(X, X^\prime)$ to specify an edit to $Y$: $\textit{Mix}(X,Y) = f(X, X', Y)$, where $X' = f(A, A', X)$.
See the supplementary material for more details.

\noindent
\textbf{Animation Transfer}: 
\modelname can also be used to create animated sequences of edited patterns. 
By leveraging parametric \dslname programs, users can generate animations for simple patterns and then apply these animations to complex patterns with no additional effort. See the video in the supplementary material for examples.

\begin{table}[t]
\small
\begin{center}
\begin{tabular}{lcccc}
\toprule
& {\footnotesize DSIM ($\downarrow$)} & {\footnotesize DISTS ($\downarrow$)} & {\footnotesize LPIPS ($\downarrow$)}& {\footnotesize SSIM ($\uparrow$)}\\
\midrule
\modelname & \textbf{0.074} & \textbf{0.184} & \textbf{0.304} & \textbf{0.704} \\
\midrule
 { - Pos. Enc.}& 0.147 & 0.239 & 0.383 & 0.659 \\
 { - Lower} &0.087 & 0.196 & 0.335 & 0.652 \\
 { - Mix} & 0.098 & 0.210 & 0.345 & 0.682\\
\midrule
Base~\cite{versatile_diffusion} & 0.585 & 0.460 & 0.815 & 0.435\\
\bottomrule
\end{tabular}
\end{center}

\caption{
Subtractive ablation study on \modelname shows that removing any component (see Section~\ref{subsec:model}) degrades performance, and that removing all components (Base) results in a sharp decline.
}
\label{tab:model_ablation}
\end{table}


\section{Conclusion}
\label{sec:conclusion}

In this paper, we introduced a novel approach for \textit{programmatic} editing of visual patterns without inferring the underlying program. By using analogies to \textit{express} desired edits and a learned conditional generative model to \textit{execute} them, our method provides an intuitive solution for pattern manipulation.
A key component of our approach is \dslname, a domain-specific language for generating diverse, structured pattern data. Paired with our procedure for sampling analogical quartets, \dslname enables the creation of a large, high-quality dataset for training. We also presented \modelname, a Latent Diffusion Model (LDM) designed to overcome critical issues that emerge when LDMs are naively deployed for analogical pattern editing, enabling high-fidelity edits that capture user intentions.
Our experiments demonstrate that \modelname successfully edits real-world patterns and surpasses baseline methods, while also generalizing to novel pattern styles beyond its training distribution. We believe that our DSL, dataset, and model will help drive further research on in-the-wild pattern image editing.
Looking forward, we aim to extend this analogical editing framework to other domains such as semi-parametric 3D modeling while continuing to improve synthetic data quality and scalability. 

{
    \small
    \bibliographystyle{ieeenat_fullname}
    \bibliography{main}
}


\end{document}